\documentclass[letterpaper, 10 pt, conference]{ieeeconf}  

\IEEEoverridecommandlockouts                              

\usepackage{graphics} 
\usepackage{amsmath} 
\usepackage{amssymb}  

\title{\LARGE \bf \approach: Self-Supervised Learning of Rotation-Equivariant Keypoint Detection and Invariant Description for Endoscopy}
\author{Mert Asim Karaoglu$^{1,2}$, Viktoria Markova$^{2}$, Nassir Navab$^{1,3}$, Benjamin Busam$^{1}$, and Alexander Ladikos$^{2}$
\thanks{$^{1}$Mert Asim Karaoglu, Nassir Navab, Benjamin Busam are with Technical University of Munich, Munich, Germany {\tt\small mert.karaoglu@tum.de}}%
\thanks{$^{2}$Mert Asim Karaoglu, Viktoria Markova, Alexander Ladikos are with ImFusion GmbH, Munich, Germany}%
\thanks{$^{3}$Nassir Navab is with Johns Hopkins University, Baltimore, MD, USA}%
}

\usepackage{multirow}
\usepackage{xspace}
\def\approach{RIDE\xspace}
\usepackage{etoolbox}
\usepackage{siunitx}
\usepackage{eurosym,rotating,booktabs}
\begin{document}

\maketitle
\thispagestyle{empty}
\pagestyle{empty}

\begin{abstract}

Unlike in natural images, in endoscopy there is no clear notion of an up-right camera orientation. Endoscopic videos therefore often contain large rotational motions, which require keypoint detection and description algorithms to be robust to these conditions.
While most classical methods achieve rotation-equivariant detection and invariant description by design, many learning-based approaches learn to be robust only up to a certain degree. At the same time learning-based methods under moderate rotations often outperform classical approaches. In order to address this shortcoming, in this paper we propose RIDE, a learning-based method for rotation-equivariant detection and invariant description.
Following recent advancements in group-equivariant learning, RIDE models rotation-equivariance implicitly within its architecture.
Trained in a self-supervised manner on a large curation of endoscopic images, RIDE requires no manual labeling of training data. We test RIDE in the context of surgical tissue tracking on the SuPeR dataset as well as in the context of relative pose estimation on a repurposed version of the SCARED dataset. In addition we perform explicit studies showing its robustness to large rotations. Our comparison against recent learning-based and classical approaches shows that RIDE sets a new state-of-the-art performance on matching and relative pose estimation tasks and scores competitively on surgical tissue tracking.

\end{abstract}


\section{INTRODUCTION}
Minimally invasive endoscopic surgery has emerged as a modern alternative to traditional open surgery, reducing patient trauma and recovery times.
During such surgeries, an endoscope is used to provide visual guidance and surveying to the operator.
However, these devices usually have certain physical drawbacks affecting maneuverability, and their limited view can make navigation difficult.
Modern computer vision techniques can be used to provide real-time solutions for simultaneous localization and mapping (SLAM)~\cite{liu2022sage, oliva2023orb, rodriguez2022tracking, rodriguez2023nr}, and tissue tracking~\cite{schmidt2022fast, schmidt2022recurrent, schmidt2023sendd}, thereby assisting surgeons in performing surgeries with more precision~\cite{fu2021future}.
Furthermore, 3D reconstruction applications~\cite{busam2018markerless, liu2020reconstructing, karaoglu2021adversarial} can be employed for diagnosis and longitudinal assessment procedures.
Detecting and describing keypoints is a crucial step in such geometric computer vision tasks.
However, due to illumination-inconsistencies and large rotational viewpoint changes, this task is exceptionally difficult for endoscopic scenes.
\begin{figure}[t!]
     \centering
     \includegraphics[width=1.0\columnwidth]{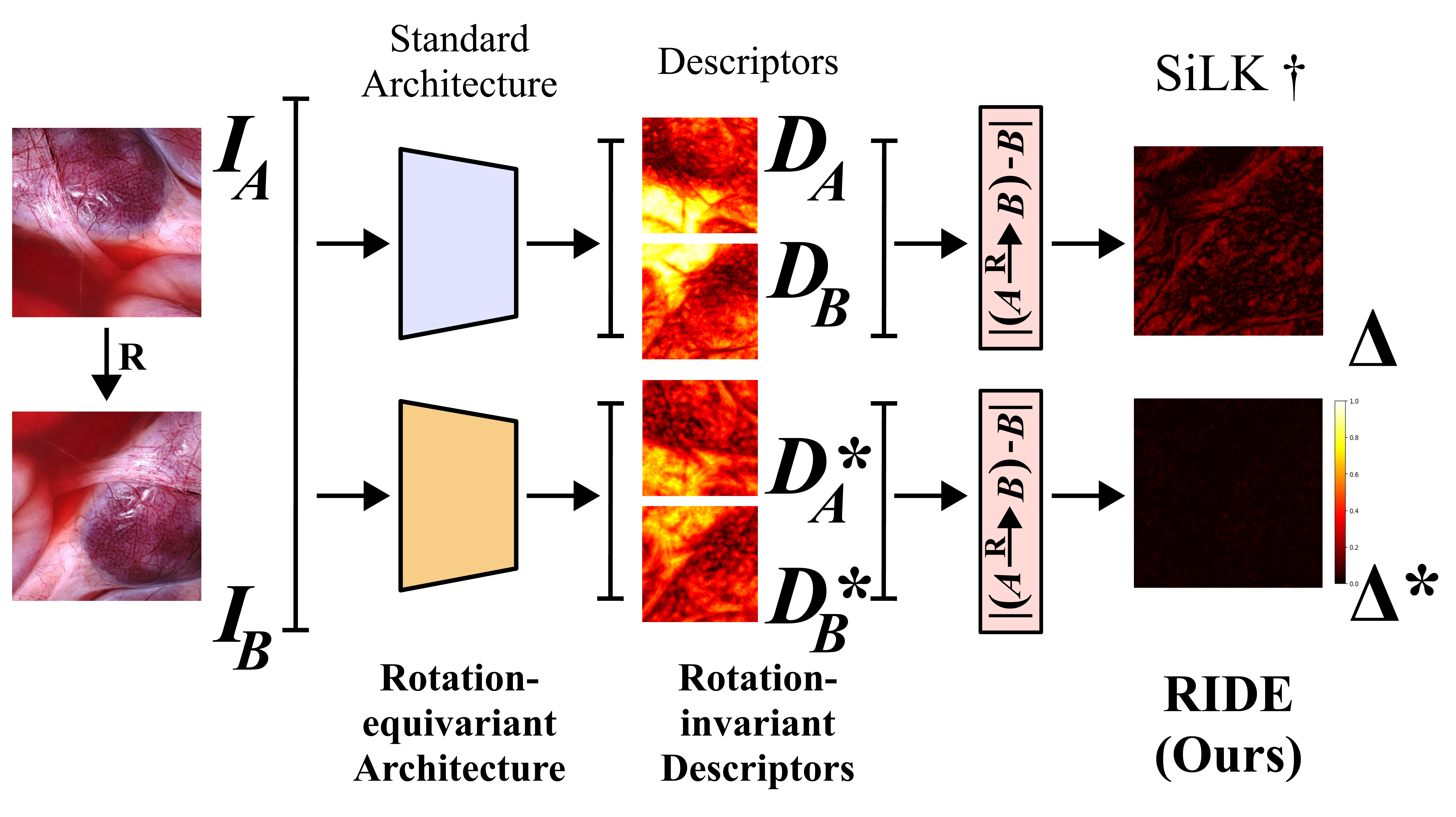}    
     \caption{In comparison to the state-of-the-art in keypoint detection and description, SiLK~\cite{gleize2023silk}, ($\dagger$ means that it is trained on endoscopic images); RIDE uses a rotation-equivariant architecture and creates rotation-invariant descriptors. Let $I_B$ be the  image $I_A$ rotated by $R$ and $D_A$, $D_B$ the respective dense descriptor maps with their dimensionalities reduced by PCA and $\Delta$ is their difference after re-alignment. We use "*" to denote our results.}
     \label{fig:introduction}
\end{figure}
Classical methods like SIFT~\cite{lowe2004distinctive} compute rotation-invariant descriptors based on estimated orientations of the keypoints.
Although they perform effectively on natural images, both their detection and matching performance substantially reduces when used for endoscopic images.
This is mainly because of the strongly non-uniform illumination, non-Lambertian surface properties, limited visual textures and frequent occlusions that exist in the environment.
More recently developed methods using convolutional neural networks (CNN) show significant improvements in detection and distinctiveness on visually demanding scenes, by learning to be robust against such inconveniences as well as viewpoint changes.
However, as we show in our experiments, their success significantly decreases, and goes below that of classical methods when challenged by large rotational motions highlighting the limitations of learning-based robustness to rotations.

In this paper, we argue for the necessity of a rotation-equivariant design for reliable keypoint detection and description on endoscopic scenes, see Fig.~\ref{fig:introduction}.
As a solution, we propose a novel method that we name RIDE.
We design RIDE substantially influenced by some of the most recent state-of-the-art works in the field, guided by the challenges of the targeted domain.
RIDE employs a rotation-equivariant~\cite{weiler2019general} steerable CNN to predict rotation-equivariant keypoints and invariant descriptors.
Its lightweight architecture makes it a great option for possible real-time applications for navigation and 3D reconstruction.
We use a simple yet effective self-supervised training scheme on homogpraphically augmented images from a large collection of endoscopic datasets requiring no manual labeling.
We extensively evaluate our method for relative pose estimation and surgical tissue tracking.
In addition we test its reliability under large rotation changes through a matching task.

In summary, we contribute:
\begin{itemize}
    \item A novel, self-supervised rotation-equivariant keypoint detection and invariant description method for endoscopic scenes with real-time capability.
    \item State-of-the-art results for endoscopic matching and relative pose estimation on the repurposed SCARED dataset~\cite{allan2021stereo}
    \item Competitive results for surgical tissue tracking on the SuPeR dataset~\cite{li2020super}.
\end{itemize}

\section{RELATED WORK}
Keypoint detection and description is one of the oldest tasks in computer vision.
Traditional methods such as SIFT~\cite{lowe2004distinctive}, ORB~\cite{rublee2011orb}, and AKAZE~\cite{alcantarilla2011fast, alcantarilla2012kaze} often remain highly competitive due to their elegant architectures, which account for various symmetries including rotation-invariance by design.
They employ various algorithms like histogram of gradient orientations~\cite{lowe2004distinctive} to compute the dominant orientations of the keypoints and use them to invariantize the descriptors.
Even though these methods can achieve remarkable results in certain uses cases, in visually challenging endoscopic images they fall behind more recent learning based approaches.

Recently, learning-based methods~\cite{detone2018superpoint, dusmanu2019d2, revaud2019r2d2, barroso2020hdd, zhao2023aliked, gleize2023silk} have proven their advantages on demanding benchmarks on natural scenes.
This success is also reflected in the surgical domain.
Liu et al.~\cite{liu2020extremely} propose a dense descriptor method for sinus endoscopy.
Similarly, ReTRo~\cite{schmidt2021real} introduces a dense descriptor model for endoscopic images and also learns to predict dense orientation maps.
Both of these methods rely on external keypoint extractors for sparse matching.
Proposing an alternative to classical methods which typically detect considerably lower number of keypoints on retinal images, GLAMPoints~\cite{truong2019glampoints} introduces a learning-based detector and uses classical descriptors for matching.
Barbed et al.~\cite{barbed2022superpoint} study the performance of SuperPoint~\cite{barbed2022superpoint}, which was designed for natural scenes, on colonoscopy images.
Their study highlights the domain-gap that negatively effects a direct transfer of such models and proposes an adaption scheme.

Similar to equivariant designs of the classical methods, recent advancements in group-equivariant learning~\cite{cohen2016group, cohen2016steerable, weiler2019general} enable constructing CNNs that are rotation-equivariant by design.
REKD~\cite{lee2022self} utilizes a rotation-equivariant steerable CNN~\cite{weiler2019general} to learn oriented keypoints using a histogram based orientation estimation inspired by SIFT~\cite{lowe2004distinctive}.
RELF~\cite{lee2023learning} employs a similar idea for rotation-invariant dense description.
Both approaches show the advantages of rotation-equivariant learning for either detection or description under large rotational motion. However, they only focus on one part of the problem, keypoint detection or description, and do not propose a joint solution.
\section{Method}
\begin{figure}[t]
    \centering
    \includegraphics[width=1.0\columnwidth]{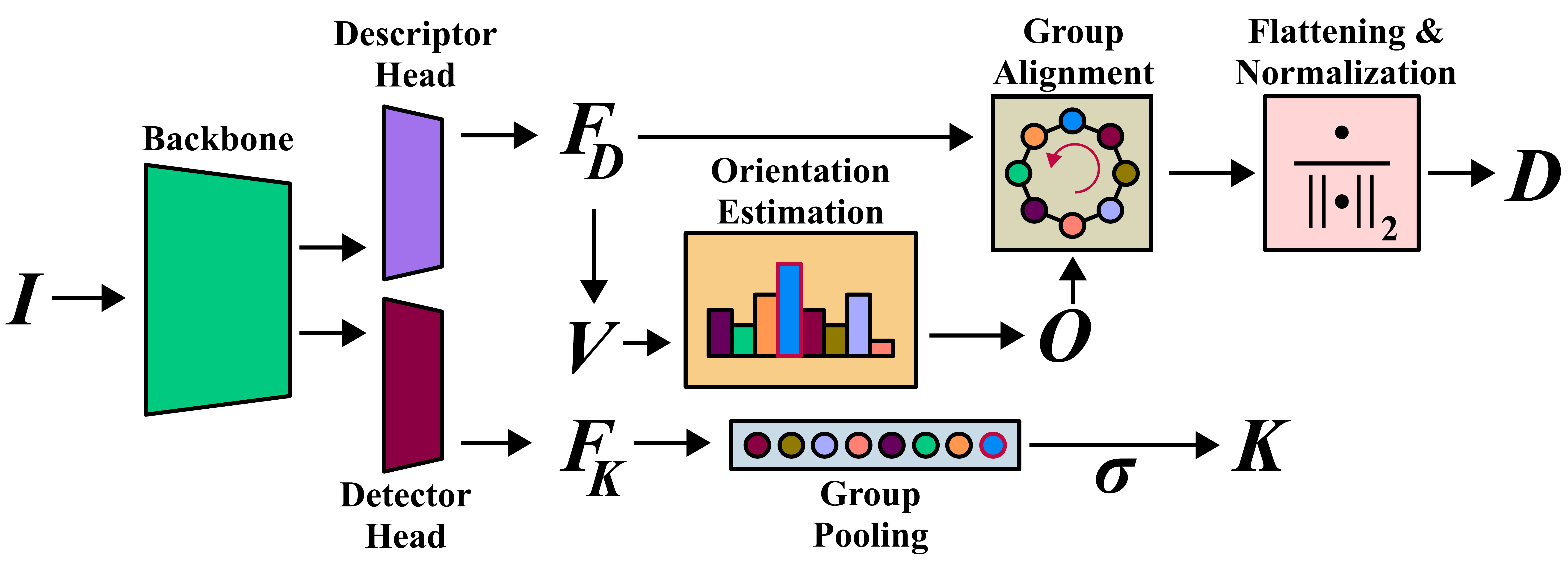}
    \caption{\textbf{RIDE's architecture}. We utilize a rotation-equivariant architecture to extract dense maps $F_D$ and $F_K$. $F_D$ is used to estimate the orientation and generate invariant descriptors via group alignment. $F_K$ is used to produce rotation-equivariant keypoint detections via group pooling.}
    \label{fig:architecture}
\end{figure}

This work proposes a real-time capable keypoint detection and description learning pipeline that can handle strong illumination changes and abrupt camera motions in endoscopic scenes.
For this, RIDE utilizes a rotation-equivariant steerable CNN and jointly learns to detect rotation-equivariant keypoints and invariant descriptors in a real-time capable architecture, see Fig.~\ref{fig:architecture}.
Our model is trained in a self-supervised manner, which allows it to learn on various endoscopic datasets without any need for manual labeling.
In the following sections, we'll explain in more detail the rotation-equivariant architecture, the keypoint detection, and the feature description components of RIDE.

\subsection{RIDE's Architecture}
Following SuperPoint~\cite{detone2018superpoint} and SiLK~\cite{gleize2023silk} we use a single model that does detection and description in two separate heads with a shared backbone.
Our model is constructed with rotation-equivariant steerable convolutions~\cite{weiler2019general}, inside a VGG-style architecture, that act on cyclic group $C_{|G|}$, $|G|$ stands for the order of the discrete group.
Even though steerable convolutions come with a small overhead cost in the training time, at inference time, they act like standard convolutions not affecting the leanness of our approach.
The detector and descriptor heads have compact architectures and output dense maps in the regular representation of the action group $G$; respectively, $F_K \in \mathbb{R}^{|G| \times H' \times W'}$ and $F_D \in \mathbb{R}^{C_D \times |G| \times H' \times W'}$, where $H'$ and $W'$ are the height and width of the maps and $C_D$ is the size of the channel dimension.

\subsection{Orientation Estimation and Rotation-invariant Description}
RIDE's descriptor head serves for both dense orientation and description estimation.
Feature map $F_D \in \mathbb{R}^{C_D \times |G| \times H' \times W'}$ contains $C_D$ number of features for each element of the action group $G$.

We follow the prior works~\cite{lee2021self, lee2022self, lee2023learning} and consider its first element along the channel dimension as the dense orientation histogram map $V \in \mathbb{R}^{|G| \times H' \times W'}$.
The indices of the histogram directly relate to the elements of the cyclic group $C_{ \left| G \right| }$ that samples the continuous rotational space $SO \left( 2 \right)$ with $\frac{360}{\left|G\right|}$ increments.
During training we extract the relative in-plane rotation from the known homography that transforms an image, $I_A$, to its warped version $I_B$.
Then, we discretize them to their nearest neighbors on the cyclic group and assign them to be the ground-truth orientations $\tilde{O_B^A}$.
We define the orientation loss term based on the histogram alignment loss introduced by Lee et al.~\cite{lee2021self}.
In our implementation, we first apply a softmax on $V$ along the group dimension.
Then, we use a cyclic shift operator that works along the group dimension, defined as $\tau_G \left(X, O \right)$, to shift the elements of the histogram $V_B$ and align to $V_A$ using the ground-truth orientations $~\tilde{O_B^A}$ as:
\begin{equation}
\begin{split}
V_A' & = \text{softmax}(V_A)_G, \\
V_B' & = \tau_G(\text{softmax}(V_B)_G, -\tilde{O_B^A}).
\end{split}
\end{equation}
Our orientation loss is defined as follows:
\begin{equation}
\mathcal{L}_O = -\frac{1}{|\tilde{M}||G|} \sum_{(i_A, i_B) \in \tilde{M}} \sum_{k=1}^{|G|} V_A'(i_A)_k \log(V_{B}'(i_B)_{k}),
\end{equation}
where $\tilde{M}$ is the list of ground-truth corresponding-pixel locations as $i_A$ on $I_A$ and $i_B$ on $I_B$.

To generate rotation-invariant descriptors, we employ group aligment~\cite{lee2023learning} and apply the cyclic shift operator $\tau_G(X, O)$ on the dense feature map $F_D$.
Similar to histogram alignment, applied on $F_D$, $\tau_G$ shifts the feature elements along the group dimension by the index of their orientations $O$ in the cyclic group $C_{|G|}$.
Then, we concatenate them along the channel dimension to create our rotation-invariant descriptors $D \in \mathbb{R}^{(C_D|G|) \times H' \times W'}$.
Finally, we normalize them to fit onto a unit-hypersphere.

For training, we use the dense descriptors, $D_A$ and $D_B$ of the image pairs and construct a dense score matrix $S$, where $S(i, j) = D_A(i) \cdot D_B(j)$.
Following ~\cite{rocco2018neighbourhood, sun2021loftr, lee2023learning}, we apply dual-softmax on $S$ and apply a temperature value to acquire the soft mutual matching probabilities $P$.
We apply negative log-likelihood on ground-truth matching pairs $\tilde{M}$ as our description loss:
\begin{equation}
\mathcal{L}_D = -\frac{1}{|\tilde{M}|} \sum_{(i_A, i_B) \in \tilde{M}} \log(P(i_A, i_B)).
\end{equation}
Please refer to our implementation and training details section for more information how $\tilde{M}$ is generated.

During inference, we use the orientation estimated from the histogram while for training we use the ground-truth orientations.
Furthmore, we leave the user the option to choose the matching algorithm, i.e. mutual nearest neighbor (MNN) or dual-softmax.

\subsection{Rotation-equivariant Keypoints}
In RIDE, we use a keypoint detection scheme constructed with simplicity in mind.
We omit using a cell-based detection which imposes spatial constraints on keypoint locations.
Instead, we follow~\cite{truong2019glampoints, gleize2023silk} and do a pixel-wise classification.

We generate the rotation-equivariant keypoint score map, $K \in \mathbb{R}^{H' \times W'}$, by collapsing the group dimension of the detector head output $F_K$ using group pooling and applying a sigmoid on it.

Following ~\cite{truong2019glampoints, tyszkiewicz2020disk, gleize2023silk}, we take the cyclic matching success of the descriptors to generate the ground-truth labels.
To generate ground-truth keypoint labels, we apply MNN between $D_A$ and $D_B$.
If ground-truth correspondences $(i_A,i_B)$ are correctly matched, we label them as keypoints on the ground-truth keypoint label maps $\tilde{K}_A$, $\tilde{K}_B$.
We use binary cross entropy (BCE) for the computing the keypoint loss as:
\begin{equation}
\begin{split}
\mathcal{L}_K =-\sum_{i \in \{A,B\}} \frac{1}{|K_i|} & \sum_{j=1}^{|K_i|} (\tilde{K}_i(j) \log(K_i(j)) \\
& + (1 - \tilde{K_i}(j)) \log(1 - K_i(j)))
\end{split}
\end{equation}
In our experiments, we do not use non-maximum suppression (NMS), but we leave it to the user's choice.

\subsection{Training Objective}
Our training loss is the combination of the orientation, description and keypoint losses:
\begin{equation}
\mathcal{L} = \lambda_O \mathcal{L}_O + \mathcal{L}_D + \mathcal{L}_K,
\end{equation}
where $\lambda_O$ is defined as the weighting factor of the orientation loss.
\section{Evaluation}
\subsection{Implementation and Training Details}
RIDE is implemented in PyTorch~\cite{paszke2017automatic} using e2cnn~\cite{weiler2019general} for group-equivariant operations.

For the experiments we train two different variations: RIDE and RIDE-L.
Even though they both follow the structure of the VGGnp-4 backbone~\cite{gleize2023silk} and the corresponding detection and description heads, RIDE-L has twice as many parameters in the channel dimension and outputs 256 dimensional descriptors.
In comparison RIDE's descriptors are 128 dimensional.
We modify this architecture by removing the bias term in the convolutions, and swapping the positions of the BatchNorm and ReLU layers so that BatchNorm is applied first.
Keeping the size of the channel dimensions the same while increasing the order of the cyclic group ($|G|$) results in high computational cost.
Based on the findings of ~\cite{han2021redet} we decide to set $|G|$ to 8 as an optimal point between runtime efficiency and performance.
Because $3 \times 3$ kernels defined on $C_8$ are not well represented when steered by 45 degrees, we employ $5 \times 5$ kernels on all convolutions~\cite{weiler2019general}.
We change the size of the channel dimensions in RIDE so that at each layer the combined size of the group and channel dimensions is equal to its corresponding layer's channel size in SiLK.
Like ~\cite{detone2018superpoint, gleize2023silk} our model also operates on grayscale images.
Since we don't use pooling functions and padding in convolutions, the size of the output maps are equal to the input size center-cropped during convolutions.
Specifically, for an input image $I$ of size $H \times W$, the output size $H' \times W'$ equal to $H-36 \times W-36$.

We train our models on a curation of various endopcopic datasets bundled and shared by Batic et al.~\cite{batic2023whether} in addition to MITI~\cite{hartwig2022miti}.
More specifically, our training set includes 179,132 images of laparoscopic operations on various anatomies, from MITI~\cite{hartwig2022miti}, DSAD~\cite{carstens2023dresden}, ESAD~\cite{bawa2020esad}, GLENDA~\cite{leibetseder2019glenda}, LapGyn4~\cite{leibetseder2018lapgyn4}, and PSI-AVA~\cite{valderrama2022towards} datasets.

We train RIDE on image pairs generated by applying known homographies similar to~\cite{detone2018superpoint, gleize2023silk}.
Taking SiLK~\cite{gleize2023silk} as the baseline, we apply the same data augmentation strategy.
This means that the angle of in-plane rotations for homography generation is limited to the range $[-22.34,22.34]$ in degrees.
We rely on the image augmentations to gain robustness to illumination-inconsistencies.
We compute the pixel correspondences, $\tilde{M}$, using the known homogprahic transformations.
Similar to SiLK~\cite{gleize2023silk} we keep only the bijective correspondences and define their positions at the pixel centers.

We train RIDE for 100,000 iterations on the curated datasets.
To balance the training, at every sample, we randomly pick one of the datasets and a frame from it.
We crop out all GUI elements visible in the images and then resize them to 480 pixels on the longest dimension.
To achieve the equal output size as SiLK, we train our models on cropped images of size $182 \times 182$.
We use the ADAM optimizer~\cite{kingma2014adam} with learning rate 1e-4 and (0.9, 0.999) as betas.
Empirically found, we set the weight for the orientation loss $\lambda_O$ to 10.
During training we apply a temperature of $20^{-1}$ on the score matrix $S$.
We train with a batch size of 2 on a single Nvidia RTX 3090 GPU with mixed precision and use the block-size computation of the score matrix $S$~\cite{gleize2023silk} to decrease the vRAM cost.
Our training of RIDE takes approximately 7 hours.

To present a fair baseline, we train a SiLK model with the VGGnp-4 backbone from scratch on the same dataset of RIDE and denote it as $\text{SiLK}\dagger$.
For all the trained models (including $\text{SiLK}\dagger$), we use the weights of the checkpoint that performed the best on the validation set.
In all the experiments we use the exact same models and extract the top 10,000 detections as keypoints.

In the experiments we use both dual-softmax and mutual nearest neighbor (MNN) matching.
For dual-softmax we always use a temperature value of 0.1 for the score matrix and matching threshold of 0.9.
These parameters are empirically chosen without detailed testing, therefore we believe that there is room for improvement for task-specific-tuning.
\subsection{Relative Pose Estimation}
\begin{figure*}[ht!]
    \centering
    \includegraphics[width=1.0\textwidth]{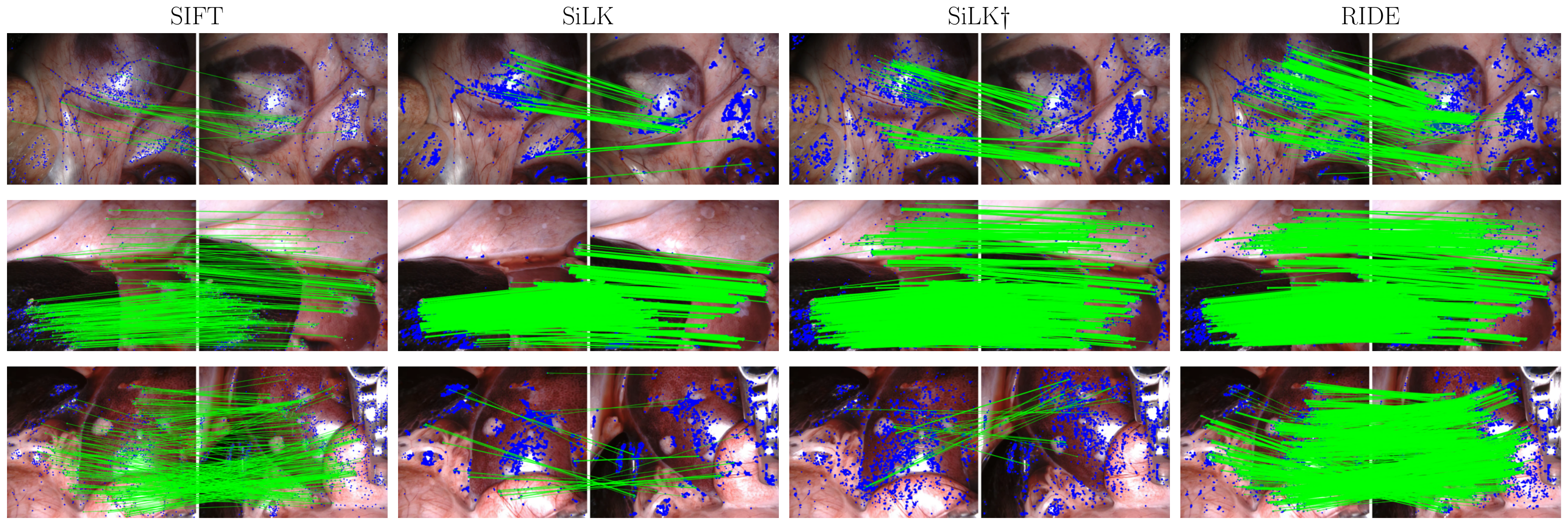}
    \caption{Qualitative matching results on pairs from SCARED dataset~\cite{allan2021stereo}. Blue points show all the detected keypoints and green lines refer to inlier matches extracting using MNN.}
    \label{fig:qualitative}
\end{figure*}
\subsubsection{Dataset}
In accordance with the assessment methods employed in previous research~\cite{sarlin2020superglue, sun2021loftr} on natural scenes~\cite{dai2017scannet}, we leverage the components from a porcine endoscopic stereo depth estimation dataset, SCARED~\cite{allan2021stereo}, to create an evaluation protocol specifically for this purpose.
For every 40\textsuperscript{th} labeled image in the training and test keyframe sequences, we iterate over the following frames in the temporal dimension and sample the pair if their underlying point clouds have an overlap between $40\%-95\%$.
We discard the pair if the source or the target image has less than $50\%$ point cloud overlap with the initial frame of the keyframe sequence.
This way, we generate in total 1,223 image pairs with challenging illumination and pose variations, see Fig.~\ref{fig:qualitative}.
We believe our sampling heuristics are more suitable for the evaluation of the task at hand in comparison to sampling based on temporal distance~\cite{barbed2022superpoint} or among left-right stereo pairs~\cite{schmidt2021real}.
In this experiment, we resize the images to $640 \times 512$.
\subsubsection{Metrics}
For evaluation we compute the AUC of the pose error under thresholds at $(5^\circ,10^\circ,20^\circ)$ as introduced by Yi et al.~\cite{yi2016lift}. The pose error is defined as the maximum of the angular error computed from the estimated rotation matrix and the unit-scale translation vector.
The relative poses are extracted from the essential matrix computed using the OpenCV~\cite{opencv_library} implementation with MAGSAC++~\cite{barath2020magsac++}.
\subsubsection{Baselines}
In addition to $\text{SiLK}\dagger$ we also compare our model against the publicly shared weights of SiLK~\cite{gleize2023silk} with the same architecture trained by the authors on the COCO dataset~\cite{lin2014microsoft}; this model is simply denoted as SiLK~\cite{gleize2023silk}.
Also we benchmark against the OpenCV~\cite{opencv_library} implementations of some of the very successful traditional methods: ORB~\cite{rublee2011orb}, AKAZE~\cite{alcantarilla2011fast, alcantarilla2012kaze}, and SIFT~\cite{lowe2004distinctive} with and without the ratio-test.
\subsubsection{Results}
As shown in Table~\ref{tab:scared_relative_pose}, both RIDE-L and RIDE achieve first and second best performances on the thresholds when coupled with the dual-softmax matcher.
We believe that the improvement shown by RIDE-L is connected to the more capacity it has due to its larger channel size.
Trailing RIDE on most metrics, $\text{SiLK}\dagger$, trained on endoscopic images, significantly improves over SiLK~\cite{gleize2023silk} trained on COCO~\cite{lin2014microsoft}.
This suggests that the domain-gap between the real-world and the endoscopic scenes is too large for learning-based methods to generalize to and therefore training, or fine-tuning on the endoscopic scenes is necessary.
The results also highlight that, unlike the other classical descriptors, SIFT can still perform respectively well under challenging conditions surpassing the SoTA learning-based algorithm, SiLK~\cite{gleize2023silk}, trained on out-of-domain data.
However, the significantly higher number of keypoint detections, and matches (see Fig.~\ref{fig:qualitative}), make learning-based approaches great options for SLAM and 3D reconstruction pipelines.
\begin{table}[h!]
    \caption{Relative pose estimation evaluation on the repurposed SCARED dataset\cite{allan2021stereo}. The most right two columns show the number of detected and matched keypoints. \textbf{Bold} values are the highest in their category and \underline{underlined} ones are the runner-ups.}
    \centering
    \resizebox{1.0\columnwidth}{!}{
    \begin{tabular}{llccccc}  \hline \hline
         & & \multicolumn{3}{c}{Pose Error AUC at} & & \\ \cline{3-5}
        Method & Matching & $5^\circ$ & $10^\circ$ & $20^\circ$ & \# Detected & \# Matched \\ \hline
        ORB~\cite{rublee2011orb} & MNN & 0.70 & 2.46 & 6.92 & 500 & 157 \\ \hline
        AKAZE~\cite{alcantarilla2011fast, alcantarilla2012kaze} & MNN & 3.22 & 11.23 & 24.48 & 443 & 172 \\ \hline
        \multirow{2}{*}{SIFT~\cite{lowe2004distinctive}} & MNN & 6.14 & 20.12 & 39.13 & 1,312 & 506 \\
        & Ratio test & 9.79 & 27.51 & 48.94 & 1,312 & 202 \\ \hline
        \multirow{2}{*}{SiLK~\cite{gleize2023silk}} & MNN & 5.17 & 16.05 & 31.21 & 10,000 & 3,183 \\
        & Dual-softmax & 8.28 & 22.47 & 40.48 & 10,000 & 1,395 \\ \hline
        \multirow{2}{*}{$\text{SiLK}\dagger$~\cite{gleize2023silk}} & MNN & 8.14 & 23.16 & 41.29 & 10,000 & 3,274 \\
        & Dual-softmax & 10.56 & 28.68 & 47.86 & 10,000 & 1,238 \\ \hline
        \multirow{2}{*}{RIDE} & MNN & 7.15 & 21.71 & 41.58 & 10,000 & 3,135 \\
        & Dual-softmax & \underline{11.03} & \underline{31.19} & \underline{54.43} & 10,000 & 1,136 \\ \hline
        \multirow{2}{*}{RIDE-L} & MNN & 8.87 & 26.26 & 48.00 & 10,000 & 3,046 \\
        & Dual-softmax & \textbf{12.05} & \textbf{33.98} & \textbf{57.84} & 10,000 & 1,273 \\ \hline\hline  
    \end{tabular}
    }
    \label{tab:scared_relative_pose}
\end{table}
\subsection{Matching Under Large In-plane Rotations}
\subsubsection{Dataset}
Unlike most real-life applications like autonomous driving or indoor navigation, in endoscopy it is common to have large-in plane rotations as part of abrupt viewpoint changes.
Since it is very challenging to get pixel-wise correspondences on endoscopic images, we create a test setup by applying known rotations.

We extract 10 temporally equally distanced images from all images in the SCARED dataset~\cite{allan2021stereo} to increase the visual variation.
Similar to the rotation invariance experiment conducted in RELF \cite{lee2023learning}, pairs of images are generated by taking the original images as the source images and their in-plane rotated version as the targets.
For each source image, the target images are its rotated version with 10 degrees increments from 0 to 350 degrees.
In total, we end-up with 360 image pairs with known pixel-wise correspondences.
We ensure that the image transformation preserves the image content scale, which generates empty parts in the resulting image. These are replaced with a smooth grayscale background.
Accordingly, while the source images are kept at $640\times512$, the target image sizes depend on the rotation angle.
\subsubsection{Metrics}
For evaluation we compute the mean matching accuracy of the matching error under thresholds at $(3,5,10)$px following~\cite{lee2023learning}.
For a fair comparison, all the methods employ MNN matching.
\subsubsection{Baselines}
We use the same baselines as the previous experiment.
\subsubsection{Results}
\begin{figure}[h!]
    \centering
    \includegraphics[width=1.0\columnwidth]{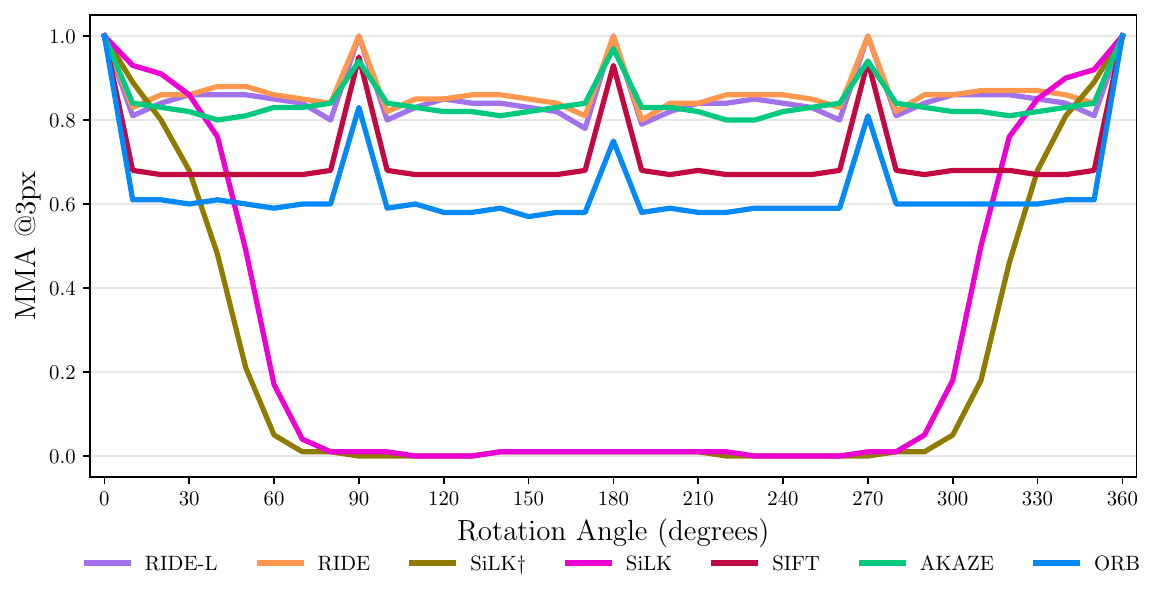}
    \caption{Mean matching accuracy (MMA) computed at the threshold of 3px. The pairs are generated using images from the SCARED dataset~\cite{allan2021stereo} and applying known rotations with 10 degree increments.}
    \label{fig:scared_360_plot}
\end{figure}

By combining strong distinctiveness and description capabilities of the learning-based methods with a rotation-invariant design that is often found in classical approaches, RIDE outperforms baselines of both classes.
As shown in Table~\ref{tab:scared_360_quantitative} and Fig.~\ref{fig:scared_360_plot}, both RIDE and RIDE-L achieve the top performance on all thresholds.
Employing a standard CNN architecture, that is only translation-equviariant, SiLK and $\text{SiLK}\dagger$ perform better than the classical methods only within a limited range of rotations.
Outside of this range, they both fail and drastically fall behind the classical methods that are engineered to be rotation invariant.
This finding strongly supports the design choices of our approach that combines the strengths of both sides.
\begin{figure*}[th]
    \centering
    \includegraphics[width=\textwidth]{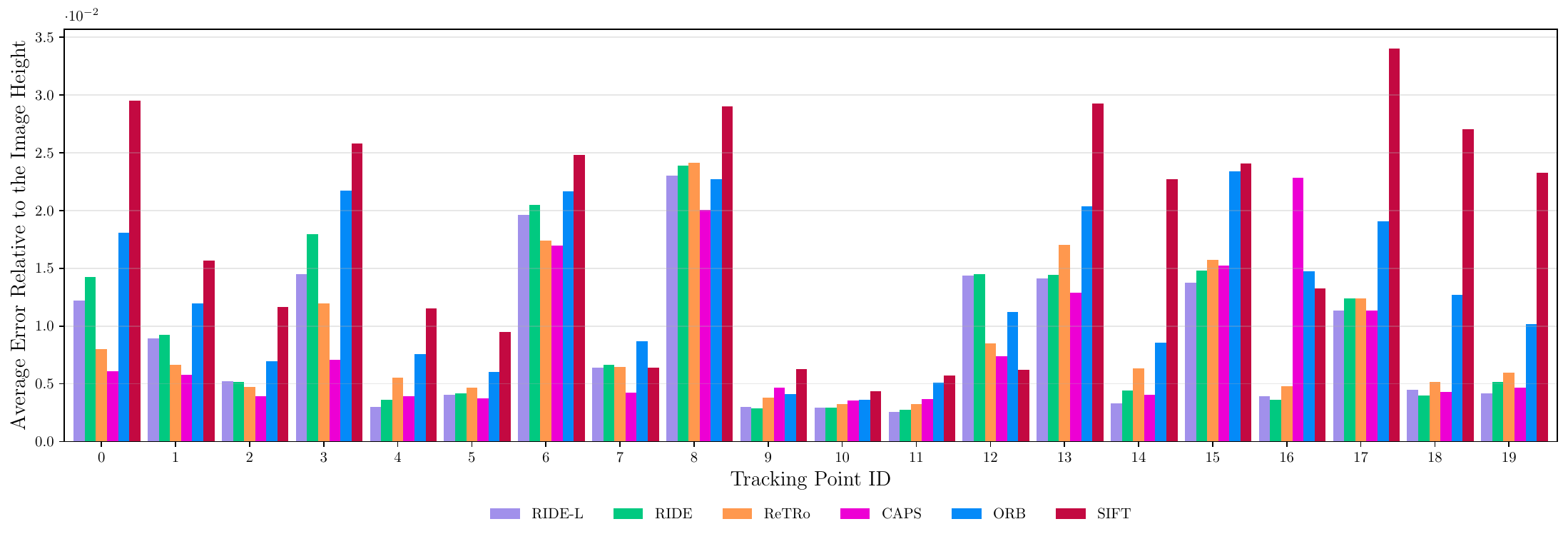}
    \caption{Average point tracking errors relative to the image height on the SuPeR dataset~\cite{li2020super}. Results of the other methods are taken from the report of Schmidt et al.~\cite{schmidt2021real}.}
    \label{fig:super}
\end{figure*}
\begin{table}[h!]
    \caption{Rotation robustness evaluation on images from SCARED dataset~\cite{allan2021stereo}. $\epsilon$ represents the threshold values in pixels used for MMA. \textbf{Bold} values are the highest in their category and \underline{underlined} ones are the runner-ups.}
    \centering
    \resizebox{0.75\columnwidth}{!}{
    \begin{tabular}{lccc}  \hline \hline
         & \multicolumn{3}{c}{Mean Matching Accuracy} \\  \cline{2-4}
        Method & $\epsilon=3$px & $\epsilon=5$px & $\epsilon=10$px \\ \hline
        ORB~\cite{rublee2011orb} & 0.62 & 0.66 & 0.68 \\ \hline
        AKAZE~\cite{alcantarilla2011fast, alcantarilla2012kaze} & 0.84 & \underline{0.85} & \underline{0.86} \\ \hline
        SIFT~\cite{lowe2004distinctive} & 0.71 & 0.71 & 0.72 \\ \hline
        SiLK~\cite{gleize2023silk} & 0.26 & 0.27 & 0.28 \\ \hline
        $\text{SiLK}\dagger$~\cite{gleize2023silk} & 0.20 & 0.21 & 0.22 \\ \hline
        RIDE & \textbf{0.87} & \textbf{0.88} & \textbf{0.89} \\ \hline
        RIDE-L & \underline{0.85} & \textbf{0.88} & \textbf{0.89} \\ \hline\hline
    \end{tabular}
    }
    \label{tab:scared_360_quantitative}
\end{table}

\subsection{Surgical Tissue Tracking}
\subsubsection{Dataset}
In this experiment we follow ReTRo's~\cite{schmidt2021real} assessment for tracking deforming points on the SuPeR dataset~\cite{li2020super}.
SuPeR~\cite{li2020super} is a perception dataset recorded on a Da Vinci Surgical System using tissue-like material to imitate a robotic surgical scene for various tasks.
Its tissue tracking subset is a sequence of 522 frames with every 10\textsuperscript{th} frame manually labeled with the tracking information of 20 pre-defined points.
Each frame is of size $640\times480$.
\subsubsection{Metrics}
We report the average distance error proportion to image height for each tracking point computed as follows.
For each labeled frame, keypoints are extracted and matched against the initial labeled frame.
From matching keypoints in the initial frame, the closest 4 keypoints are assigned to a tracking point.
Their average motion is computed and recorded as the motion of the tracking point.
And finally, for each tracking point, the average distance error in pixels divided by the image height is reported.
\subsubsection{Baselines}
In this experiment we compare against ReTRO~\cite{schmidt2021real}, CAPS~\cite{wang2020learning}, ORB~\cite{rublee2011orb} and SIFT~\cite{lowe2004distinctive}.
Their results are taken from the report of Schmidt et al.~\cite{schmidt2021real}.
For RIDE and RIDE-L we compute the matches with dual-softmax and following~\cite{schmidt2021real}, we employ the OpenCV~\cite{opencv_library} implementation of the GMS~\cite{Bian2020gms}.
\subsubsection{Results}
Based on the results depicted in the Fig.~\ref{fig:super}, our model demonstrates competitive performance compared to the learning-based methods~\cite{wang2020learning, schmidt2021real} exceeding well above the classical approaches~\cite{oliva2023orb, lowe2004distinctive}.
This shows the robustness of our method even on deforming tissues, which is a transformation it does not encounter during training.
\subsection{Runtime Performance}
Compiled with TensorRT on half precision, for an input of size $640 \times 512$ RIDE runs at 65.41 FPS while the larger model RIDE-L achieves 17.91 FPS on average on an NVIDIA GeForce RTX 3090.
\section{DISCUSSION}
In this work, we mainly focus on solving the task of reliable keypoint detection and description for endoscopic images that contain large rotational viewpoint changes.
On the task of relative pose estimation, see Table~\ref{tab:scared_relative_pose} and Fig.~\ref{fig:qualitative}, our method excels over the classical approaches and the state-of-the-art CNN-based method~\cite{gleize2023silk} proving the effectiveness of its rotation-equivariant architecture on providing reliable predictions regardless of the large viewpoint changes.
This experiment also highlights that the domain-gap between natural and surgical images is too large for learning-based methods trained on the prior to generalize successfully.
Moreover, our method's robustness is further challenged in the matching task showing consistent success across the whole spectrum of rotation angles making it robust to rotations beyond those seen during training.
Finally, we test our method for deforming tissue tracking and perform competitively to the state-of-the-art learning-based method for endoscopy~\cite{schmidt2021real}.

However, endoscopic videos consist of many more challenges other than large rotational viewpoint changes.
Our method relies on image augmentations for learning to be robust against illumination-inconsistencies.
Yet, as it is displayed in Fig.~\ref{fig:qualitative}, RIDE also detects keypoints on the edges of reflections which can result in adversities.
In future work, regularizing terms~\cite{barbed2022superpoint} can be studied and imposed on the detection objective.
Furthermore, tissue deformation is a prominent issue in surgeries. A more sophisticated architecture directly targeting deformation-awareness~\cite{potje2023cvpr} can be further explored in the context of endoscopy.
\section{CONCLUSION}
We present RIDE, a self-supervised rotation-equivariant detection and invariant description method for endoscopic scenes.
As a learning-based method, RIDE can predict distinctive descriptors and high numbers of salient keypoints.
Furthermore, its rotation-equivariant design enables it to perform reliably under large rotational motion.
RIDE achieves state-of-the-art performance on matching and relative pose estimation tasks and scores competitively on surgical tissue tracking, outperforming recent learning-based and classical approaches.







\section*{Acknowledgment}{%
We thank Adam Schmidt for providing us with great support in reproducing the surgical tissue tracking experiment.%
}

\bibliographystyle{IEEEtran} 
\bibliography{IEEEabrv, root} 

\end{document}